\documentclass[11pt]{article}

\usepackage[final]{acl}

\usepackage{times}
\usepackage{latexsym}

\usepackage[T1]{fontenc}

\usepackage[utf8]{inputenc}

\usepackage{microtype}

\usepackage{graphicx}
\usepackage{amsmath}
\usepackage{booktabs}

\title{XTurnix: Large-Scale Self-Supervised Turn Control\\
  through Two-State Binary Decisions}

\author{Zhanxun Liu\textsuperscript{1,2,3}\thanks{Primary contributor.}
  \And\hspace{-0.42in} Yifan Duan\textsuperscript{1,2}
  \And\hspace{-0.68in} Hengtao Wu\textsuperscript{1}
  \And\hspace{-0.68in} Chen Yang\textsuperscript{2,4}
  \And\hspace{-0.68in} Qinyuan Cheng\textsuperscript{2,4}
  \AND\vspace{-0.12in}
  Kun Wang\textsuperscript{5}
  \And\hspace{-0.68in} Xingyu Zeng\textsuperscript{6}
  \And\hspace{-0.68in} Xipeng Qiu\textsuperscript{2,4}
  \And\hspace{-0.68in} Chaochao Lu\textsuperscript{3}
  \And\hspace{-0.68in} Xie Chen\textsuperscript{1,2}\thanks{Corresponding author.}
  \AND\vspace{-0.12in}
  \parbox{0.95\linewidth}{\centering\normalfont
    \textsuperscript{1}MoE Key Lab of Artificial Intelligence, X-LANCE Lab, Shanghai Jiao Tong University,\\
    \textsuperscript{2}Shanghai Innovation Institute,\quad
    \textsuperscript{3}Shanghai AI Laboratory,\quad
    \textsuperscript{4}MOSI Intelligence,\\
    \textsuperscript{5}SenseTime Group Inc.,\quad
    \textsuperscript{6}Shenzhen University of Advanced Technology\\
    \href{https://github.com/xcc-zach/xturnix}{Code}\quad
    \href{https://huggingface.co/spaces/xcczach/xturnix-demo}{Demo}}
}

\newcommand{\listening}{\texttt{<|listening|>}}
\newcommand{\speaking}{\texttt{<|speaking|>}}
\newcommand{\startaction}{\texttt{<|start|>}}
\newcommand{\keepaction}{\texttt{<|keep|>}}
\newcommand{\stopaction}{\texttt{<|stop|>}}
\newcommand{\pausetoken}{\texttt{<|pause|>}}

\begin{document}
\maketitle

\begin{abstract}
General turn-taking behavior in real-time dialogue systems requires deciding
whether to keep listening or start responding while listening, and whether to
continue or stop while speaking. Existing turn detectors use heterogeneous,
task-specific label spaces and are often trained on limited annotations or
evaluated on isolated utterances, making them difficult to use as a unified
causal controller with comprehensive context. We propose XTurnix, a compact
text-based model that formulates turn control as two binary decisions
conditioned on the AI's current listening or speaking state and predicts a
single control token from the complete dialogue history. XTurnix is pretrained
on 5.5 million causal action examples automatically derived from timestamped
two-speaker transcripts, then fine-tuned on synthetic multi-turn examples with
a flatter distribution across the four state-action labels. We evaluate
XTurnix on four public benchmarks and a balanced
self-curated benchmark. Across the public benchmarks, XTurnix leads on nearly every benchmark: it
is best on all SemanticVAD and LiveKit splits, ties the native Smart-Turn
model on Smart-Turn bench, and achieves the highest incomplete-turn accuracy
on Easy-Turn. On the self-curated benchmark, it reaches 89.06\% accuracy,
more than 20 points above the strongest third-party baseline (68.75\%), while
keeping F1 scores between 84.21\% and 90.63\% across all four categories. These results demonstrate unified listening- and speaking-state
turn control in a single compact model.
\end{abstract}

\section{Introduction}\label{sec:introduction}

\begin{figure*}[t]
  \centering
  \includegraphics[width=\textwidth]{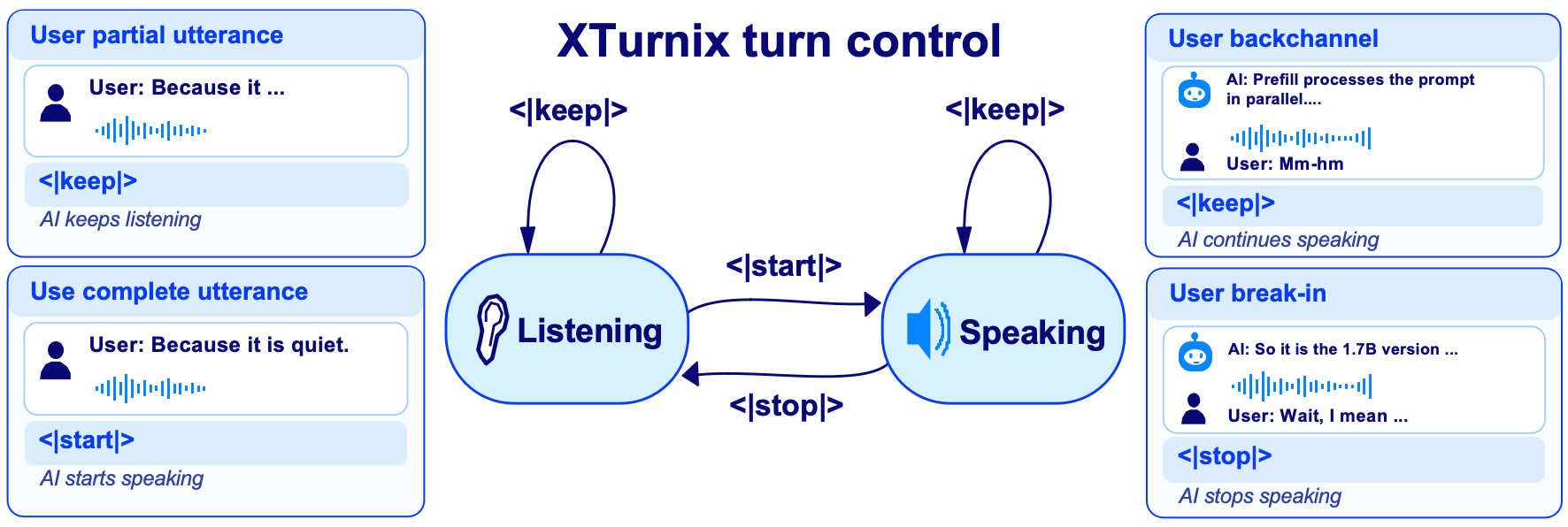}
  \caption{XTurnix overview}
  \label{fig:overview}
\end{figure*}

A real-time conversational AI must continuously manage the conversational
floor. While listening, it should begin speaking only after the user has
yielded the floor; starting too early interrupts an unfinished utterance, while
starting too late creates an unnatural pause. While speaking, it must
distinguish a short acknowledgment that permits the current response to
continue from an interruption that requires it to stop and listen. We refer to
this state-conditioned control problem as \emph{turn detection}.

Existing audio- and text-based turn detectors address utterance completion,
backchannel handling, and multilingual endpoint detection. Their input
modalities and output categories vary, however, making it difficult to use
them through a common turn-control interface
(Section~\ref{sec:related-work}).

A general turn controller must address three challenges. First, it needs an
action space that directly expresses whether to keep or transfer the floor,
conditioned on whether the AI is listening or speaking. Second, it needs
scalable training supervision; timestamped conversation corpora offer an
opportunity to derive turn decisions without manual task-specific labels.
Third, it must interpret incremental user input using the full dialogue
history. Earlier questions, instructions, and commitments can all affect
whether the same fragment calls for a response, continued listening, or an
interruption.

As shown in Figure \ref{fig:overview},we address these challenges with XTurnix, a compact text model that predicts a
single turn-control token from the complete dialogue history and the AI's
current state. The state makes the legal action space explicit: while
listening, the model chooses whether to keep listening or start speaking; while
speaking, it chooses whether to keep speaking or stop. Text input allows
incremental decisions from automatic speech recognition (ASR) output
while retaining multi-turn context.

Our contributions are threefold:
\begin{itemize}
  \item We develop a self-supervised procedure that temporally interleaves
    tokens from timestamped two-speaker transcripts and derives token-level
    action labels through a state machine, enabling large-scale pretraining
    without manual turn annotations.
  \item We formulate turn detection as two state-conditioned binary
    classification problems, each with two legal actions. Three control
    actions, with \keepaction{} shared across states, express all four logical
    turn behaviors without conflating actions with semantic event types.
  \item We instantiate this approach in XTurnix, a compact text-based turn
    controller, and evaluate it on four public Chinese benchmark subsets and
    a balanced multi-turn benchmark. XTurnix achieves competitive performance
    on public benchmarks and the highest overall accuracy among the evaluated
    models on the multi-turn benchmark, with strong F1 scores across all four
    state-action categories.
\end{itemize}

\section{Related Work}\label{sec:related-work}

TurnGPT
\citep{ekstedt-skantze-2020-turngpt} predicts, as a dialogue unfolds, whether
the current speaker has reached a point where another speaker could take
over. It uses the preceding dialogue to distinguish a complete response from
an unfinished fragment. Voice Activity Projection (VAP)
\citep{ekstedt2022vap} predicts when each participant will speak over the next
two seconds. It learns from speech activity in recorded conversations. These
predictions indicate whether the current speaker will continue, the other
participant will take over, or a brief backchannel will occur.
DualTurn \citep{rajaa2026dualturn} uses generative pretraining on dual-channel
conversational audio to support turn-taking prediction and interaction
control.
The BERT-based endpoint module in FireRedChat
\citep{chen2025fireredchat} classifies accumulated ASR transcripts, while
Namo Turn Detector v1 Multilingual \citep{videosdk2025namo} uses an
mmBERT-based classifier for multilingual complete/incomplete text detection.
Smart-Turn \citep{pipecat2026smartturn} estimates completion from audio with
a lightweight Whisper-based classifier, typically in conjunction with voice
activity detection. Easy-Turn \citep{li2025easyturn} combines a Whisper
encoder with a language model to generate transcriptions and classify speech
as complete, incomplete, backchannel, or wait.
SoulX-Duplug \citep{yan2026soulxduplug} combines streaming audio
representations with ASR-derived text to predict five user states:
idle, non-idle, complete, incomplete, and backchannel.
It processes speech in 160\,ms chunks to support semantic endpoint detection
and interruption handling in full-duplex spoken dialogue.
X2-Turn \citep{fu2026x2turn}
builds on Voxtral Realtime \citep{mistralai2026voxtralrealtime} to jointly
transcribe streaming speech and predict turn states. Separate ASR and
turn-state heads share the same streaming representations and produce
frame-level predictions. TurnSense
\citep{baiji2026turnsense} classifies audio as complete, incomplete, or invalid,
supporting semantic endpoint detection and invalid-input filtering. TEN Turn
Detection \citep{ten2025turndetection} uses Qwen2.5-7B to classify text as
finished, unfinished, or wait, with the latter covering requests to pause or
terminate the interaction.
Anyreach's Semantic Turn-Taking model
\citep{rajaa2026semanticturntaking} adapts Qwen2.5-0.5B-Instruct using synthetic
conversations to predict turn-taking actions from dialogue context.

\section{Task Formulation}\label{sec:task-formulation}

\begin{table*}[t]
  \centering
  \small
  \begin{tabular}{lp{0.48\linewidth}ll}
    \toprule
    \textbf{Current AI state} & \textbf{Interpretation of the latest user input}
      & \textbf{AI Action} & \textbf{Next AI state} \\
    \midrule
    Listening & The user has not yielded the current turn.
      & \keepaction{} & Listening \\
    Listening & The user has yielded the turn and a response may begin.
      & \startaction & Speaking \\
    Speaking & The input is a backchannel or overlap that does not request the floor.
      & \keepaction{} & Speaking \\
    Speaking & The input is an interruption that requests the floor.
      & \stopaction & Listening \\
    \bottomrule
  \end{tabular}
  \caption{The two-state binary turn-detection formulation.}\label{tab:task-formulation}
\end{table*}

We define four state-conditioned labels by pairing each of the two dialogue
states with its two legal control actions, as summarized in
Table~\ref{tab:task-formulation}.
Let $m_i=(r_i,x_i)$ denote the $i$-th dialogue message, where
$r_i\in\{\mathrm{user},\mathrm{AI}\}$ is the speaker role and $x_i$ is its
textual content. Let $H_t=(m_1,\ldots,m_{n_t})$ be the dialogue history
available at decision point $t$, where $n_t$ is the number of messages in the
history. The final message is always a user message, so
$r_{n_t}=\mathrm{user}$. As streaming ASR updates its content $x_{n_t}$, each
update defines a new decision point and triggers one model inference.
During deployment, an upstream input processor inserts \pausetoken{} into
the current user message when the end of a speech segment is detected by
VAD or streaming ASR.
This insertion triggers another turn-decision inference. Let
$s_t\in\{\mathrm{L},\mathrm{S}\}$ denote the AI state at that point: listening
($\mathrm{L}$) or speaking ($\mathrm{S}$). The detector is a policy
\begin{equation}
  a_t=f_\theta(H_t,s_t), \qquad a_t\in\mathcal{A}(s_t),
  \label{eq:policy}
\end{equation}
with a state-dependent binary action space
\begin{align}
  \mathcal{A}(\mathrm{L}) &= \{\keepaction{},\startaction{}\}, \\
  \mathcal{A}(\mathrm{S}) &= \{\keepaction{},\stopaction{}\}.
\end{align}

The state transition is deterministic:
\begin{equation}
  s_{t+1} =
  \begin{cases}
    \mathrm{S}, & a_t=\startaction{}, \\
    \mathrm{L}, & a_t=\stopaction{}, \\
    s_t,        & a_t=\keepaction{}.
  \end{cases}
  \label{eq:state-transition}
\end{equation}

\section{Method}\label{sec:method}

\begin{figure*}[t]
  \centering
  \includegraphics[width=0.75\textwidth]{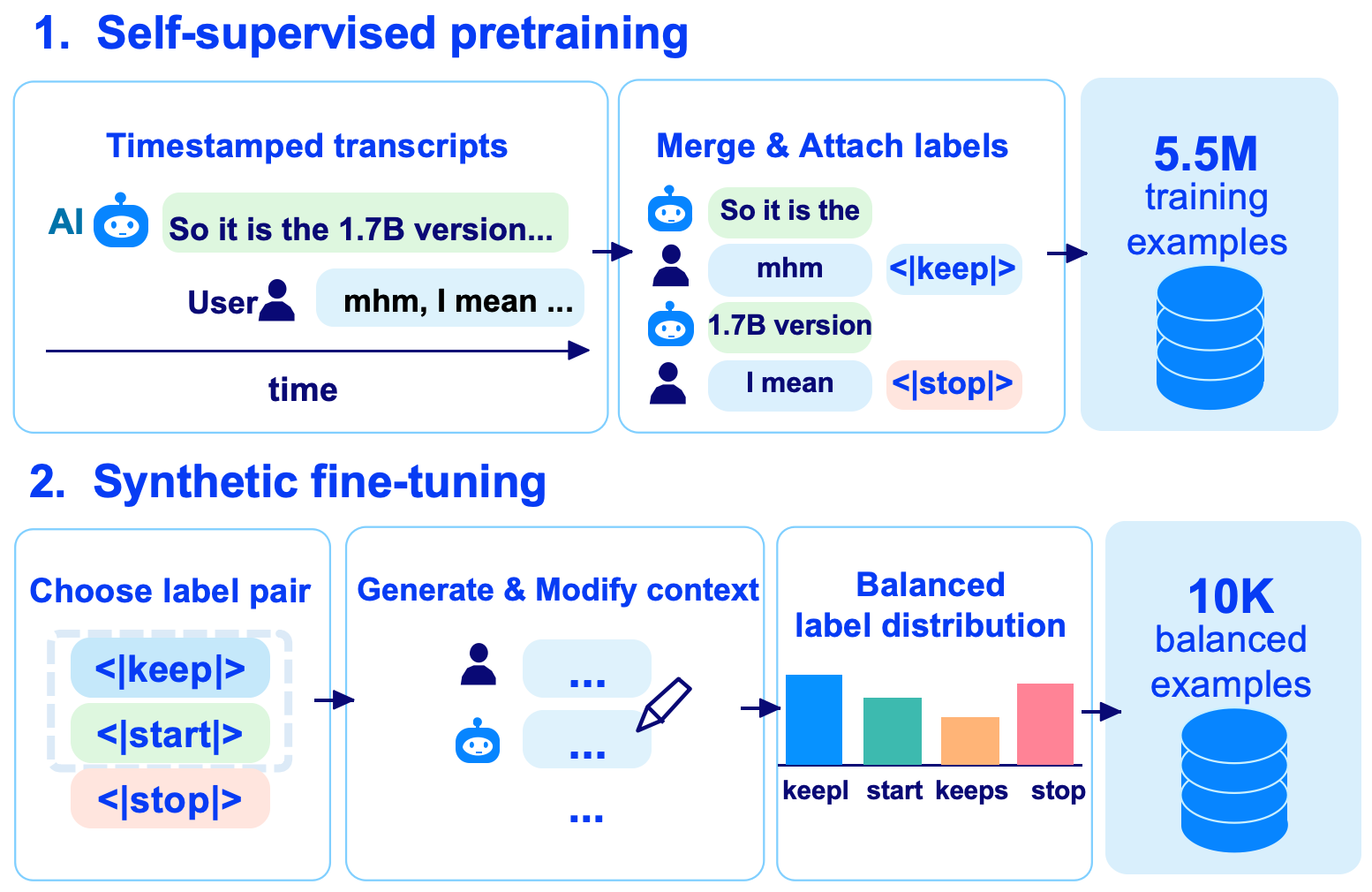}
  \caption{XTurnix training stages}
  \label{fig:training}
\end{figure*}

As shown in Figure \ref{fig:training}, XTurnix is trained in two stages. We first derive token-level supervision from
timestamped speech transcripts and perform large-scale task-specific
pretraining. We then synthesize multi-turn examples for all four state-action
pairs and fine-tune the model on the resulting, substantially flatter label
distribution. Samples of the training data are shown in
Table~\ref{tab:training-examples}.

\subsection{Self-Supervised Training Data Construction}\label{sec:self-supervised-data}

\paragraph{Source corpus.}
Our source collection contains 2,702,974 audio clips totaling 173,717.0 hours.
It includes transcripts from podcasts, films and television, animation, short
videos, sports content, audio programs, and cross-talk performances. The
collection contains Chinese, English, and a small proportion of other languages.
After filtering for valid two-speaker transcripts, we retain 689,669 clips
totaling 44,142.5 hours, with an average duration of 230.4 seconds.
The retained clips contain a mean of 23.13 turns, with a median of 12 and a
90th percentile of 52. Overlapping speech occurs in 20.18\% of clips and
accounts for 0.82\% of the time during which at least one speaker is active.
The construction below yields 280,740,536 action examples; punctuation
normalization and conflict filtering leave 276,852,309 examples in the
available pool.

\paragraph{Timestamped token timeline.}
Each source record contains a two-speaker transcript whose segments have the
form $[b][\text{speaker}]\ \text{text}\ [e]$, where $b$ and $e$ are the
segment start and end times. We retain records with exactly two speakers,
assign the speaker who appears first to the user role and the other speaker to
the assistant role, and tokenize every segment with the Qwen3 fast tokenizer
\citep{yang2025qwen3} without adding boundary tokens.

Because timestamps are provided at segment rather than token level, we assign
uniform subintervals. For a segment spanning $[b,e]$ with $n$ tokens, token
$i\in\{0,\ldots,n-1\}$ receives
\begin{equation}
  \left[
    b+\frac{i(e-b)}{n},
    b+\frac{(i+1)(e-b)}{n}
  \right].
  \label{eq:token-time}
\end{equation}
If two adjacent user segments are separated by silence, we insert the atomic
token \pausetoken{} over the gap. We do not insert pauses between assistant
segments or across speaker changes. Finally, we sort all user and assistant
tokens by start time to obtain one interleaved timeline.

\paragraph{State-machine labeling.}
We scan the merged timeline from an initial listening state. Each consecutive
run of assistant tokens is appended directly to the dialogue history of the
next training example, without creating intermediate examples. We then set the
state to speaking and retain the end time of the run's final token for labeling
subsequent user tokens. Every user token, including \pausetoken, creates one
causal example whose input contains the dialogue prefix through that token.

Labels follow directly from temporal order:
\begin{itemize}
  \item In the listening state, the target is \keepaction{} if the next token
    in the merged timeline is also a user token. Otherwise the target is
    \startaction, since the observed prefix has reached the user-to-assistant
    turn boundary.
  \item In the speaking state, let $e_A$ be the end time of the latest
    assistant token and $b_A^+$ the start time of the next assistant token.
    The target is \keepaction{} when $b_A^+$ exists and
    $b_A^+\leq e_A+0.1$ seconds, indicating that the assistant continued
    through the overlapping user input. Otherwise the target is
    \stopaction.
\end{itemize}
After emitting \startaction{} or \stopaction, the state is updated according to
Equation~\ref{eq:state-transition}. Each example consists of a system message
that declares the current state, the complete text history, and one action
token as the target.

Raw transcript conventions can introduce a shortcut: a final punctuation mark
may reveal that a user segment is complete. Before training, we therefore
strip trailing Unicode punctuation from the final user prefix. If the same
normalized input occurs with both \keepaction{} and \startaction{}, we discard
the conflicting \keepaction{} example. This prevents punctuation and duplicate
boundaries from dominating the learned decision rule.

\subsection{Synthetic Training Data Construction}\label{sec:synthetic-data}

The self-supervised corpus is highly imbalanced. Its empirical proportions
for \listening{}/\keepaction{}, \listening{}/\startaction{},
\speaking{}/\keepaction{}, and \stopaction{} are
$94.46\%$, $2.62\%$, $0.51\%$, and $2.41\%$, respectively. Direct sampling
would provide little supervision for speaking-state behavior. We therefore
adopt an \emph{output-first} procedure
\citep{wang-etal-2023-self-instruct}: choose a target state-action label before
generating or transforming its input.

Let $\mathcal{Y}$ denote the set of four state-action labels, and let $p_y$ be
the empirical proportion of label $y\in\mathcal{Y}$. We define the synthetic
target distribution as
\begin{equation}
  q_y =
  \frac{p_y^{1-\lambda}}
       {\sum_{y'\in\mathcal{Y}}p_{y'}^{1-\lambda}},
  \qquad \lambda=0.6.
  \label{eq:synthetic-distribution}
\end{equation}
This yields target proportions of $62.78\%$, $14.96\%$, $7.78\%$, and
$14.48\%$ for the four labels above. We set a target number of examples for
each label according to these proportions. During generation, we prioritize
the label that is furthest behind its target. This keeps the generated dataset
close to the target distribution throughout the process, rather than
correcting the class balance only at the end.

We sample a transcript as a topic seed and ask DeepSeek-V4-Pro-Preview
\citep{deepseekai2025deepseekv3} to produce a natural multi-turn user-assistant
context with strict role alternation. Each context contains at least two user
and two assistant messages and ends with an assistant message. We reuse a
context for exactly two distinct labels, preferring a pair from the same state.
We keep most of the dialogue context unchanged and modify only the
turn-relevant part so that the paired examples have different correct actions,
following prior work on counterfactually augmented data and contrast sets
\citep{kaushik-etal-2020-learning,gardner-etal-2020-evaluating}.

After selecting a target state-action label, we apply the corresponding
transformation to the shared context:
\begin{itemize}
  \item \textbf{Listening}/\keepaction{}: remove the final assistant response and
    truncate the preceding user message at an internal position.
  \item \textbf{Listening}/\startaction{}: remove the final assistant response but
    retain the complete preceding user message.
  \item \textbf{Speaking}/\keepaction{}: truncate the final assistant message and
    generate a short user backchannel that allows the assistant to continue
    speaking.
  \item \textbf{Speaking}/\stopaction{}: truncate the final assistant message,
    generate a user interruption with a new conversational obligation, and
    truncate that interruption to resemble incremental input.
\end{itemize}

\begin{table*}[t]
  \centering
  \small
  \begin{tabular}{llp{0.49\linewidth}p{0.18\linewidth}}
    \toprule
    \textbf{State} & \textbf{Action} & \textbf{Earlier dialogue}
      & \textbf{Latest user input} \\
    \midrule
    Listening & \keepaction{}
      & \textbf{User:} I am considering a lakeside town for my weekend
        trip.\newline
        \textbf{AI:} Why does that destination appeal to you?
      & \textbf{User:} Because it \\
    Listening & \startaction
      & \textbf{User:} I am considering a lakeside town for my weekend
        trip.\newline
        \textbf{AI:} Why does that destination appeal to you?
      & \textbf{User:} Because it is quiet \\
    Speaking & \keepaction{}
      & \textbf{User:} How does prefill differ from decoding?\newline
        \textbf{AI:} Prefill processes the prompt in parallel, whereas
        decoding generates one token at a time and uses
      & \textbf{User:} Mm-hm \\
    Speaking & \stopaction
      & \textbf{User:} I loaded the smaller checkpoint.\newline
        \textbf{AI:} So it is the 1.7B version, right? If not
      & \textbf{User:} Yes \\
    \bottomrule
  \end{tabular}
  \caption{Examples of the four training classes formed by the two current
  states and their two legal outputs. The latest user input is the decision
  point; incomplete text intentionally has no terminal punctuation.}
  \label{tab:training-examples}
\end{table*}

Trailing punctuation is removed from the last user message in every class.
Generated outputs are parsed into the same message format as the
self-supervised examples.

\subsection{Training}\label{sec:training}

We initialize XTurnix from Qwen3-0.6B \citep{yang2025qwen3} and register six
atomic special tokens:
the two state tokens \listening{} and \speaking, the three action tokens
\startaction, \keepaction, and \stopaction, and the silence token
\pausetoken. The maximum sequence length is 2,048 tokens.

We perform task-specific pretraining for 21,500 optimization steps, consuming
5,504,000 examples from the self-supervised data pool, with eight devices, a per-device batch size
of 32, and a learning rate of $2\times10^{-5}$. To reduce
the dominance of frequent labels during training, we weight each example's loss
by $\sqrt{n_{\max}/n_c}$, where $n_c$ is the number of training examples in its
state-action class and $n_{\max}$ is the size of the largest class.
We select the checkpoint at step 21,500 because it achieves the lowest
validation loss among the evaluated pretraining checkpoints. We denote
the checkpoint after this stage
\textbf{XTurnix-PT}.

Starting from the 21,500-step checkpoint, we fine-tune on 10,026
general-purpose synthetic examples for one epoch. This stage uses eight
devices with a per-device batch size of 16, a learning rate of $5\times10^{-6}$,
and completes in 79 optimization steps. No self-supervised replay examples
are mixed into this run. Both stages train the model to emit exactly one legal
action token, so inference requires neither free-form decoding nor
post-processing of natural-language output. We denote the resulting model
\textbf{XTurnix-ZH-Base}.

\begin{table*}[!t]
  \centering
  \small
  \setlength{\tabcolsep}{3pt}
  \begin{tabular}{@{}lr*{10}{c}@{}}
    \toprule
    \multicolumn{2}{c}{} & \textbf{Easy-Turn Bench} & \multicolumn{3}{c}{\textbf{Smart-Turn Bench}}
      & \multicolumn{3}{c}{\textbf{SemanticVAD Bench}}
      & \multicolumn{3}{c}{\textbf{LiveKit Bench}} \\
    \cmidrule(lr){3-3}\cmidrule(lr){4-6}\cmidrule(lr){7-9}\cmidrule(lr){10-12}
    \multicolumn{2}{c}{\textbf{Model}} & \textbf{ZH}
      & \textbf{All} & \textbf{ZH} & \textbf{EN}
      & \textbf{All} & \textbf{ZH} & \textbf{EN}
      & \textbf{All} & \textbf{ZH} & \textbf{EN} \\
    \midrule
    Easy-Turn & 850M & \textbf{97.33} & -- & 13.31 & -- & -- & 70.06 & -- & -- & 55.45 & -- \\
    Smart-Turn & 8M & 77.17 & \textbf{98.81} & \textbf{98.52} & \underline{99.00} & 54.61 & 53.09 & 69.75 & \underline{70.91} & 64.60 & \underline{76.24} \\
    FireRedChat & 170M & 81.17 & 9.29 & 1.78 & 14.34 & 72.42 & 72.59 & 70.75 & 42.67 & 42.91 & 42.48 \\
    Namo & 307M & 50.83 & 44.17 & 48.52 & 41.24 & 46.14 & 45.68 & 50.75 & 49.92 & 44.20 & 54.75 \\
    TEN & 7B & 90.33 & 74.40 & 74.85 & 74.10 & 62.65 & 61.14 & 77.75 & 63.50 & 61.20 & 65.45 \\
    TurnSense & 47M & \underline{96.17} & 79.05 & 94.97 & 68.33 & 64.86 & 64.32 & 70.25 & 58.08 & 58.73 & 57.52 \\
    SoulX-Duplug & 0.6B & 64.83 & 94.52 & 92.31 & 96.02 & 72.69 & 74.02 & 59.50 & 61.67 & 58.85 & 64.06 \\
    X2-Turn & 4B & 76.50 & \underline{98.57} & \underline{97.04} & \textbf{99.60} & 80.66 & 80.88 & \underline{78.50} & 64.79 & 59.79 & 69.01 \\
    \midrule
    Qwen3-0.6B & 0.6B & 49.50 & 0.00 & 0.00 & 0.00 & 38.33 & 37.76 & 44.00 & 43.05 & 40.45 & 45.25 \\
    Synthetic-only & 0.6B & 61.86 & 59.81 & 50.48 & 60.92 & 65.97 & 66.85 & 57.00 & 52.92 & 53.58 & 53.20 \\
    XTurnix-PT & 0.6B & 86.00 & \textbf{98.81} & \textbf{98.52} & \underline{99.00} & \underline{84.08} & \underline{84.97} & 75.25 & \textbf{73.70} & \textbf{70.57} & \textbf{76.34} \\
    XTurnix-ZH-Base & 0.6B & 95.33 & 97.38 & 96.75 & 97.81 & \textbf{92.14} & \textbf{93.36} & \textbf{80.00} & \underline{70.91} & \underline{66.47} & 74.65 \\
    \bottomrule
  \end{tabular}
  \caption{Accuracy (\%) on Chinese (ZH) and English (EN). Easy-Turn Bench excludes backchannel examples, and Smart-Turn Bench is restricted to \texttt{endfiller=true}. Best results are bold, and second-best underlined.}\label{tab:third-party-results}
\end{table*}

\begin{table*}[!t]
  \centering  
  \begin{minipage}[t]{0.40\textwidth}
    \centering
    \scriptsize
    \setlength{\tabcolsep}{5pt}
    \begin{tabular}{@{}lrr@{}}
      \toprule
      \textbf{Model} & \multicolumn{2}{c}{\texttt{endfiller}} \\
      \cmidrule(lr){2-3}
        & \textbf{False} & \textbf{True} \\
      \midrule
      Easy-Turn & 82.23 & 13.31 \\
      Smart-Turn & \textbf{85.96} & \textbf{98.52} \\
      FireRedChat & 80.71 & 1.78 \\
      Namo & 53.81 & 48.52 \\
      TEN & \underline{84.43} & 74.85 \\
      TurnSense & 62.77 & 94.97 \\
      SoulX-Duplug & 44.50 & 92.31 \\
      X2-Turn & 43.32 & \underline{97.04} \\
      \midrule
      Qwen3-0.6B & 80.03 & 0.00 \\
      XTurnix-PT & 31.47 & \textbf{98.52} \\
      XTurnix-ZH-Base & 58.88 & 96.75 \\
      \bottomrule
    \end{tabular}
    \caption{Accuracy (\%) on Smart-Turn Bench (ZH) grouped by the
      \texttt{endfiller} metadata field. Best results are bold, and second-best
      results underlined.}
    \label{tab:smartturn-endfiller}
  \end{minipage}%
  \hfill
  \begin{minipage}[t]{0.56\textwidth}
    \centering
    \scriptsize
    \setlength{\tabcolsep}{3.5pt}
    \begin{tabular}{@{}lrr@{}}
      \toprule
      \textbf{Model} & \textbf{Complete} & \textbf{Incomplete}
        \\
      \midrule
      Easy-Turn & \underline{96.33} & 98.33 \\
      Smart-Turn & 87.00 & 67.33 \\
      FireRedChat & 95.00 & 67.33 \\
      Namo & 64.67 & 37.00 \\
      TEN & 89.33 & 91.33 \\
      TurnSense & 95.00 & 97.33 \\
      SoulX-Duplug & 35.67 & 94.00 \\
      X2-Turn & 57.33 & 95.67 \\
      \midrule
      Qwen3-0.6B & \textbf{99.00} & 0.00 \\
      XTurnix-PT & 69.00 & \textbf{100.00} \\
      XTurnix-ZH-Base & 91.33 & \underline{99.33} \\
      \bottomrule
    \end{tabular}
    \caption{Class-specific accuracy (\%) on Easy-Turn Bench. Best results are
      bold, and second-best results underlined.}
    \label{tab:easyturn-breakdown}
  \end{minipage}
\end{table*}

\begin{table*}[!t]
  \centering
  \scriptsize
  \setlength{\tabcolsep}{4pt}
  \begin{tabular}{@{}llrrrr@{}}
    \toprule
    \textbf{Evaluation slice} & \textbf{Model} & \textbf{ROC-AUC}
      & \startaction{} \textbf{precision} & \startaction{} \textbf{recall}
      & \keepaction{} \textbf{recall} \\
    \midrule
    Easy-Turn endpoint & Qwen3-0.6B & 35.34 & 49.75 & 99.00 & 0.00 \\
     & XTurnix-PT & 99.28 & 100.00 & 69.00 & 100.00 \\
     & XTurnix-ZH-Base & 99.28 & 99.28 & 91.33 & 99.33 \\
    \addlinespace
    Smart-Turn, \texttt{endfiller=false} & Qwen3-0.6B & 57.88 & 80.03 & 100.00 & 0.00 \\
     & XTurnix-PT & 87.92 & 94.74 & 15.22 & 96.61 \\
     & XTurnix-ZH-Base & 87.36 & 94.92 & 51.37 & 88.98 \\
    \bottomrule
  \end{tabular}
  \caption{Turn-completion results (\%)
    on Easy-Turn endpoint examples and Smart-Turn (ZH) examples with
    \texttt{endfiller=false}. ROC-AUC measures how well the model separates
    \startaction{} from \keepaction{} across all thresholds. Easy-Turn
    results exclude backchannels.}
  \label{tab:xturnix-start-conservatism}
\end{table*}

\begin{table*}[!t]
  \centering
  \small
  \begin{tabular}{@{}lrrrrr@{}}
    \toprule
    & \textbf{Overall} & \multicolumn{2}{c}{\textbf{Listening F1}}
      & \multicolumn{2}{c}{\textbf{Speaking F1}} \\
    \cmidrule(lr){3-4}\cmidrule(lr){5-6}
    \textbf{Model} & \textbf{Accuracy} & \keepaction{} & \startaction{}
      & \keepaction{} & \stopaction{} \\
    \midrule
    Easy-Turn & 57.03 & 65.31 & 78.48 & 38.71 & 42.42 \\
    Smart-Turn & 47.66 & 61.22 & 75.95 & 11.11 & 35.14 \\
    FireRedChat & 61.72 & 0.00 & 66.67 & 69.09 & 76.71 \\
    Namo & 46.88 & 36.36 & 52.05 & 44.07 & 52.17 \\
    TEN & 51.56 & \textbf{93.55} & \textbf{93.94} & 3.33 & 14.71 \\
    TurnSense & 55.47 & 88.57 & 86.21 & 34.67 & 7.55 \\
    SoulX-Duplug & 57.81 & 78.48 & 65.31 & 9.76 & 57.47 \\
    X2-Turn & 68.75 & 76.06 & 70.18 & 53.06 & 70.89 \\
    \midrule
    Qwen3-0.6B & 51.56 & 55.88 & 50.00 & 66.67 & 0.00 \\
    XTurnix-PT & \underline{86.72} & 88.89 & 85.71 & \textbf{87.32} & \underline{84.21} \\
    XTurnix-ZH-Base & \textbf{89.06} & \underline{90.63} & \underline{90.63} & \underline{86.21} & \textbf{88.57} \\
    \bottomrule
  \end{tabular}
  \caption{Overall accuracy and class-specific F1 scores (\%) on the
    self-curated benchmark. Best results in each column are bold, and
    second-best results underlined.}\label{tab:benchmark-v1-results}
\end{table*}

\section{Experiments}\label{sec:experiments}

\subsection{Evaluation Benchmarks}\label{sec:evaluation-benchmarks}

\paragraph{Third-party benchmarks.}
We evaluate on four public turn-detection benchmarks.
Easy-Turn contains 700 complete, incomplete, and backchannel utterances
\citep{li2025easyturn}. Smart-Turn v3.2 provides 31,527 endpoint and
non-endpoint examples \citep{pipecat2026smartturndata}. SemanticVAD contributes
4,391 examples spanning completion, incompletion, backchannel, and interruption
\citep{keteam2025semanticvad}. LiveKit EoT Bench contributes 12,471 causal
decision points labeled as end of turn or hold \citep{livekit2026eotbench}.

\paragraph{Self-Curated Benchmark.}
We also construct a small Chinese, text-only benchmark for multi-turn turn
detection. It contains 128 manually curated examples and is exactly balanced
across the four state-action combinations: 32 \listening{}/\keepaction{}, 32
\listening{}/\startaction{}, 32 \speaking{}/\keepaction{}, and 32
\speaking{}/\stopaction{} examples. Each example
provides the dialogue history, the AI's current state, and an incremental final
user message. In contrast to isolated end-of-utterance tests, correct
predictions may require interpreting the latest fragment against earlier
questions, constraints, acknowledgments, or corrections.

\subsection{Models and Evaluation Protocol}\label{sec:evaluation-protocol}

We evaluate the Easy-Turn model, Smart-Turn v3.2, FireRedChat Turn Detector,
Namo Turn Detector v1 Multilingual, TEN Turn Detection, TurnSense, X2-Turn-4B-0812,
zero-shot Qwen3-0.6B, the Synthetic-only model fine-tuned solely on 10,026
synthetic examples, XTurnix-PT, and
XTurnix-ZH-Base. The third-party models are cited as follows:
\citep{li2025easyturn,pipecat2026smartturn,chen2025fireredchat,
videosdk2025namo,ten2025turndetection,baiji2026turnsense,fu2026x2turn}.
Easy-Turn, Smart-Turn, TurnSense, and X2-Turn consume audio, whereas FireRedChat,
Namo, TEN, Qwen3-0.6B, and Synthetic-only consume text. XTurnix-PT and
XTurnix-ZH-Base are evaluated as described in Section~\ref{sec:training}.

We report accuracy on the full eligible samples of each benchmark. Each benchmark's native labels
and each model's native output probabilities are mapped to the actions in
Section~\ref{sec:task-formulation}. The AI state is assigned from the benchmark
protocol before inference. Complete or end-of-turn events map to \startaction{} while
listening; incomplete events map to \keepaction{}; backchannels map to
\keepaction{} while speaking; and interruptions map to \stopaction{}.

Models receive inputs appropriate to their modality. We use original audio
when a benchmark provides it and synthesize audio with VoxCPM2
\citep{voxcpmteam2026voxcpm2} when an audio-based detector is evaluated on
text-only data. Text models receive native transcripts when available and
transcripts generated by Qwen3-ASR-1.7B \citep{shi2026qwen3asr} otherwise. For
Smart-Turn, XTurnix additionally removes terminal Chinese and ASCII periods,
matching the punctuation normalization used in training. On the self-curated
benchmark, XTurnix receives the original multi-turn history and explicit AI
state. FireRedChat, Namo, and TEN receive only the final user input through
their native single-text interfaces, while audio models receive synthesized speech.

\subsection{Results on Third-Party Benchmarks}\label{sec:third-party-results}

Within XTurnix checkpoints, fine-tuning primarily recalibrates the model's
action preference. In the pretraining data, \startaction{} and \stopaction{}
account for only 5.03\% of the examples, compared with 29.47\% in the
fine-tuning data. After class weighting, their share of the training signal
increases from 23.31\% to 41.76\%. XTurnix-ZH-Base consequently becomes less biased toward
\keepaction{} and more likely to predict a transition action.

Table~\ref{tab:third-party-results} shows that this recalibration changes performance according to the benchmark
distribution. Relative to XTurnix-PT, XTurnix-ZH-Base improves Chinese accuracy
by 9.33\% on Easy-Turn and 8.39\% on SemanticVAD, while decreasing it by
1.77\% on Smart-Turn and 4.10\% on LiveKit. The gains and losses reflect how
the fine-tuned action preference aligns with each benchmark's class distribution.

Across models, Table~\ref{tab:third-party-results} shows that XTurnix leads on
nearly every benchmark. The XTurnix checkpoints jointly hold the best result on
all SemanticVAD and LiveKit splits: XTurnix-ZH-Base reaches 92.14\% on
SemanticVAD (All), ahead of the strongest third-party model's 80.66\%, while
XTurnix-PT reaches 73.70\% on LiveKit (All), ahead of 70.91\%. On Smart-Turn,
XTurnix-PT ties the native Smart-Turn model for the best pooled and Chinese
results (98.81\% and 98.52\%). Only on
Easy-Turn, the one benchmark that a native system wins outright, does XTurnix
not lead: the Easy-Turn model reaches 97.33\%, ahead of the best XTurnix
result of 95.33\%. The analyses below provide insight into specific
benchmarks.

Table~\ref{tab:smartturn-endfiller} indicates the importance of an \texttt{endfiller} in Smart-Turn Bench
. The
\texttt{endfiller} field indicates whether an utterance ends with a filler or
hesitation such as ``um'' or ``er''. When this cue is annotated, XTurnix-ZH-Base reaches 96.75\%, close
to Smart-Turn's 98.52\%; without it, their scores are 58.88\% and 85.96\%.
Without an explicit \texttt{endfiller} cue, turn status is more ambiguous:
even a semantically complete utterance may be interpreted either as a finished
contribution or as a premise for further elaboration. The decision therefore
depends on its discourse context and the model's action preference. For
example, Smart-Turn labels the statement ``If a weather event occurs at a
particular time of year, the available data indicate that, um, it is likely to
recur at almost the same location the following year'' as \keepaction{}, whereas
both XTurnix checkpoints predict \startaction{}. Either decision is plausible:
the statement can stand alone or introduce a subsequent explanation.

Table~\ref{tab:easyturn-breakdown} reveals the same ambiguity in Easy-Turn Bench's
complete class. Turn completeness can not be determined within only one sentence.  By contrast, incomplete examples generally contain explicit
continuation-projecting structure and offer a more determinate decision.
XTurnix-PT achieves the best incomplete accuracy across all compared models
at 100.00\%, while XTurnix-ZH-Base ranks second at 99.33\%, ahead of
Easy-Turn's 98.33\%.

Backchannel acknowledgments pose a difficulty analogous to turn
completeness. Single-token echoes such as ``mhm'' or ``right'' have no
meaningful turn interpretation in isolation: within a context, they may be a
mere backchannel or an affirmative answer to a question.
Turn detection on such utterances is therefore not well-defined, and we exclude
backchannel samples when evaluating on the Easy-Turn Bench.

Table~\ref{tab:xturnix-start-conservatism} shows how well XTurnix separates
\startaction{} from \keepaction{} and how it performs when choosing the more
probable action.
ROC-AUC measures how well the predicted \startaction{} probability ranks
completed turns above continuing turns. Both checkpoints achieve 99.28\%
ROC-AUC on Easy-Turn; on Smart-Turn \texttt{endfiller=false}, XTurnix-PT and
XTurnix-ZH-Base reach
87.92\% and 87.36\%, respectively. These high ROC-AUC values indicate
strong discrimination between completed and continuing turns. In
deployment, the decision boundary can therefore be calibrated on a small
target-domain validation set without retraining the model.

Precision and recall reveal XTurnix's conservative preference for
\keepaction{}. For both XTurnix checkpoints in Table~\ref{tab:xturnix-start-conservatism},
\startaction{} precision exceeds 94\%, meaning
that most predicted \startaction{} actions are correct. Its lower
\startaction{} recall, however, shows that many completed turns are still
classified as \keepaction{}. Fine-tuning reduces this bias: \startaction{}
recall rises from 69.00\% to 91.33\% on Easy-Turn and from 15.22\% to 51.37\%
on Smart-Turn, with only minor changes in precision. This improvement comes
with lower \keepaction{} recall, which falls from 100.00\% to 99.33\% and from
96.61\% to 88.98\%, respectively. XTurnix-ZH-Base is therefore more responsive
than XTurnix-PT
while retaining reliable \startaction{} predictions.

\subsection{Results on the Self-Curated Benchmark}\label{sec:benchmark-v1-results}

Table~\ref{tab:benchmark-v1-results} shows that XTurnix is both the strongest
overall and the most balanced across the four decisions. XTurnix-ZH-Base and
XTurnix-PT reach
89.06\% and 86.72\% accuracy, compared with 68.75\% for the strongest
third-party model. All XTurnix class F1 scores lie between 84.21\% and 90.63\%,
whereas every baseline has at least one substantially weak class. The
results therefore show that XTurnix's overall advantage is supported by
consistent performance across all four decisions rather than by one strong
class.

On the two listening decisions, TEN reaches the highest F1 scores at 93.94\%
for \startaction{} and 93.55\% for \keepaction{}, while XTurnix-ZH-Base ranks second at
90.63\% on both. XTurnix therefore remains competitive with a strong endpoint
detector even though it is trained to handle both listening and speaking
states---and it does so with a model over an order of magnitude smaller:
XTurnix builds on a 0.6B backbone, whereas TEN is a 7B-model.

On the two speaking decisions, the XTurnix checkpoints occupy the top two
positions for both \keepaction{} and \stopaction{}. Most baselines are endpoint
detectors rather than native speaking-state controllers; their speaking
predictions are projections of endpoint scores without multi-turn history or
AI-state input. XTurnix, by contrast, natively supports deciding whether to
continue or stop while speaking, giving it a broader set of turn-control
capabilities.

Within XTurnix, fine-tuning raises overall accuracy from 86.72\% to 89.06\%.
The largest class changes are on the transition actions:
\listening{}/\startaction{} F1 rises from 85.71\% to 90.63\%, and
\stopaction{} rises from 84.21\% to 88.57\%. The two \keepaction{} classes
remain close, changing from 88.89\% to 90.63\% while listening and from 87.32\%
to 86.21\% while speaking. XTurnix-ZH-Base thus improves responsiveness while preserving
balanced performance across the four decisions.

\section{Conclusion}\label{sec:conclusion}

We studied general turn-taking behavior in dialogue systems and formulated it
as two state-conditioned binary decisions for listening and speaking. We
introduced XTurnix, trained through self-supervision on large-scale timestamped
conversational data and augmented with synthetic multi-turn examples. XTurnix
achieves leading or competitive performance across third-party benchmarks and
the strongest, most balanced four-class performance on our self-curated
benchmark. These results show that a single model can decide when to start
responding, keep listening, continue speaking, or stop speaking, rather than
only detect user-turn endpoints.

\bibliography{custom}

@article{li2025easyturn,
  title = {Easy Turn: Integrating Acoustic and Linguistic Modalities for Robust
    Turn-Taking in Full-Duplex Spoken Dialogue Systems},
  author = {Li, Guojian and Wang, Chengyou and Xue, Hongfei and Wang, Shuiyuan
    and Gao, Dehui and Zhang, Zihan and Lin, Yuke and Li, Wenjie and Xiao,
    Longshuai and Fu, Zhonghua and Xie, Lei},
  journal = {arXiv preprint arXiv:2509.23938},
  year = {2025},
  url = {https://arxiv.org/abs/2509.23938}
}

@misc{pipecat2026smartturn,
  title = {{Smart-Turn v3.2}: Native Audio Turn Detection Model},
  author = {{Pipecat AI}},
  year = {2026},
  howpublished = {Software release},
  url = {https://github.com/pipecat-ai/smart-turn}
}

@misc{pipecat2026smartturndata,
  title = {{Smart-Turn v3.2} Test Dataset},
  author = {{Pipecat AI}},
  year = {2026},
  howpublished = {Hugging Face dataset},
  url = {https://huggingface.co/datasets/pipecat-ai/smart-turn-data-v3.2-test}
}

@article{chen2025fireredchat,
  title = {{FireRedChat}: A Pluggable, Full-Duplex Voice Interaction System
    with Cascaded and Semi-Cascaded Implementations},
  author = {Chen, Junjie and Hu, Yao and Li, Junjie and Li, Kangyue and Liu,
    Kun and Li, Wenpeng and Li, Xu and Li, Ziyuan and Shen, Feiyu and Tang, Xu
    and Wei, Manzhen and Wu, Yichen and Xie, Fenglong and Xu, Kaituo and Xie,
    Kun},
  journal = {arXiv preprint arXiv:2509.06502},
  year = {2025},
  url = {https://arxiv.org/abs/2509.06502}
}

@misc{videosdk2025namo,
  title = {{Namo Turn Detector v1}: Multilingual},
  author = {{VideoSDK Team}},
  year = {2025},
  howpublished = {Hugging Face model},
  url = {https://huggingface.co/videosdk-live/Namo-Turn-Detector-v1-Multilingual},
  note = {ONNX-optimized mmBERT for turn detection in 23 languages}
}

@misc{ten2025turndetection,
  title = {{TEN Turn Detection}: Turn Detection for Full-Duplex Dialogue
    Communication},
  author = {{TEN Team}},
  year = {2025},
  howpublished = {Software release},
  url = {https://github.com/TEN-framework/ten-turn-detection}
}

@misc{baiji2026turnsense,
  title = {{TurnSense and TurnSense 1.1}: Three-Class Semantic Turn Detection
    for Chinese and English Speech Interaction},
  author = {{Baiji Team}},
  year = {2026},
  howpublished = {Hugging Face model},
  url = {https://huggingface.co/brgroup/TurnSense}
}

@article{yang2025qwen3,
  title = {{Qwen3} Technical Report},
  author = {Yang, An and Li, Anfeng and Yang, Baosong and Zhang, Beichen and
    Hui, Binyuan and Zheng, Bo and Yu, Bowen and Gao, Chang and Huang, Chengen
    and Lv, Chenxu and others},
  journal = {arXiv preprint arXiv:2505.09388},
  year = {2025},
  url = {https://arxiv.org/abs/2505.09388}
}

@misc{keteam2025semanticvad,
  title = {{SemanticVAD} Dialogue State Detection Dataset},
  author = {{KE-Team}},
  year = {2025},
  howpublished = {Hugging Face dataset},
  url = {https://huggingface.co/datasets/KE-Team/SemanticVAD-Dataset}
}

@misc{livekit2026eotbench,
  title = {{eot-bench}: The Open Benchmark for End-of-Turn Detection},
  author = {{LiveKit}},
  year = {2026},
  howpublished = {Software and dataset benchmark},
  url = {https://github.com/livekit/eot-bench}
}

@misc{voxcpmteam2026voxcpm2,
  title = {{VoxCPM2}: Tokenizer-Free {TTS} for Multilingual Speech Generation,
    Creative Voice Design, and True-to-Life Cloning},
  author = {{VoxCPM Team}},
  year = {2026},
  howpublished = {Software and model release},
  url = {https://github.com/OpenBMB/VoxCPM}
}

@inproceedings{wang-etal-2023-self-instruct,
  title = {Self-Instruct: Aligning Language Models with Self-Generated
    Instructions},
  author = {Wang, Yizhong and Kordi, Yeganeh and Mishra, Swaroop and Liu,
    Alisa and Smith, Noah A. and Khashabi, Daniel and Hajishirzi, Hannaneh},
  booktitle = {Proceedings of the 61st Annual Meeting of the Association for
    Computational Linguistics (Volume 1: Long Papers)},
  year = {2023},
  address = {Toronto, Canada},
  publisher = {Association for Computational Linguistics},
  pages = {13484--13508},
  doi = {10.18653/v1/2023.acl-long.754},
  url = {https://aclanthology.org/2023.acl-long.754/}
}

@article{deepseekai2025deepseekv3,
  title = {{DeepSeek-V3} Technical Report},
  author = {{DeepSeek-AI} and Liu, Aixin and Feng, Bei and Xue, Bing and Wang,
    Bingxuan and others},
  journal = {arXiv preprint arXiv:2412.19437},
  year = {2025},
  url = {https://arxiv.org/abs/2412.19437}
}

@inproceedings{kaushik-etal-2020-learning,
  title = {Learning the Difference that Makes a Difference with
    Counterfactually-Augmented Data},
  author = {Kaushik, Divyansh and Hovy, Eduard H. and Lipton, Zachary C.},
  booktitle = {International Conference on Learning Representations},
  year = {2020},
  url = {https://openreview.net/forum?id=Sklgs0NFvr}
}

@inproceedings{gardner-etal-2020-evaluating,
  title = {Evaluating Models' Local Decision Boundaries via Contrast Sets},
  author = {Gardner, Matt and Artzi, Yoav and Basmov, Victoria and Berant,
    Jonathan and Bogin, Ben and Chen, Sihao and Dasigi, Pradeep and Dua, Dheeru
    and others},
  booktitle = {Findings of the Association for Computational Linguistics:
    EMNLP 2020},
  year = {2020},
  address = {Online},
  publisher = {Association for Computational Linguistics},
  pages = {1307--1323},
  doi = {10.18653/v1/2020.findings-emnlp.117},
  url = {https://aclanthology.org/2020.findings-emnlp.117/}
}

@inproceedings{ekstedt-skantze-2020-turngpt,
  title = {{TurnGPT}: a Transformer-based Language Model for Predicting
    Turn-taking in Spoken Dialog},
  author = {Ekstedt, Erik and Skantze, Gabriel},
  booktitle = {Findings of the Association for Computational Linguistics:
    EMNLP 2020},
  year = {2020},
  address = {Online},
  publisher = {Association for Computational Linguistics},
  pages = {2981--2990},
  doi = {10.18653/v1/2020.findings-emnlp.268},
  url = {https://aclanthology.org/2020.findings-emnlp.268/}
}

@inproceedings{ekstedt2022vap,
  title = {Voice Activity Projection: Self-supervised Learning of
    Turn-taking Events},
  author = {Ekstedt, Erik and Skantze, Gabriel},
  booktitle = {Interspeech 2022},
  year = {2022},
  pages = {5190--5194},
  doi = {10.21437/Interspeech.2022-10955},
  url = {https://www.isca-archive.org/interspeech_2022/ekstedt22_interspeech.html}
}

@misc{shi2026qwen3asr,
  title = {{Qwen3-ASR Technical Report}},
  author = {{Qwen Team}},
  year = {2026},
  journal = {arXiv preprint arXiv:2601.21337},
  url = {https://arxiv.org/abs/2601.21337}
}

@misc{rajaa2026semanticturntaking,
  title = {Semantic Turn-Taking Model},
  author = {Rajaa, Shangeth},
  year = {2026},
  howpublished = {Hugging Face model card},
  url = {https://huggingface.co/anyreach-ai/semantic-turn-taking}
}

@article{rajaa2026dualturn,
  title = {{DualTurn}: Learning Turn-Taking from Dual-Channel Generative
    Speech Pretraining},
  author = {Rajaa, Shangeth},
  journal = {arXiv preprint arXiv:2603.08216},
  year = {2026},
  url = {https://arxiv.org/abs/2603.08216}
}

@article{yan2026soulxduplug,
  title = {{SoulX-Duplug}: Plug-and-Play Streaming State Prediction
    Module for Realtime Full-Duplex Speech Conversation},
  author = {Yan, Ruiqi and Chen, Wenxi and Liu, Zhanxun and Ma, Ziyang
    and Lin, Haopeng and Wen, Hanlin and Xie, Hanke and Wu, Jun
    and Liang, Yuzhe and Zhao, Yuxiang and Feng, Pengchao
    and Qian, Jiale and Meng, Hao and Dai, Yuhang and Yin, Shunshun
    and Tao, Ming and Xie, Lei and Yu, Kai and Wang, Xinsheng
    and Chen, Xie},
  journal = {arXiv preprint arXiv:2603.14877},
  year = {2026},
  url = {https://arxiv.org/abs/2603.14877}
}

@article{fu2026x2turn,
  title = {{X2-Turn}: Frame-Synchronous Dual-Head Modeling for Joint
    Streaming ASR and Turn State Prediction},
  author = {Fu, Kaiqi and Wen, Rime and Lin, Altman and Qin, Shawn
    and Gan, Roy and Wang, Hao and Wang, Qian},
  journal = {arXiv preprint arXiv:2608.10878},
  year = {2026},
  url = {https://arxiv.org/abs/2608.10878}
}

@article{mistralai2026voxtralrealtime,
  title = {{Voxtral Realtime}},
  author = {{Mistral-AI} and Liu, Alexander H. and Ehrenberg, Andy
    and Lo, Andy and Sun, Chen-Yo and Lample, Guillaume
    and Delignon, Jean-Malo and others},
  journal = {arXiv preprint arXiv:2602.11298},
  year = {2026},
  url = {https://arxiv.org/abs/2602.11298}
}

\clearpage
\appendix
\onecolumn
\section{Complementary Pooled and English Turn-Detection Results}
\label{sec:pooled-english-results}

\begin{table}[htbp]
  \centering
  \small
  \setlength{\tabcolsep}{6pt}
  \begin{tabular}{@{}lllllll@{}}
    \toprule
    & \multicolumn{3}{c}{\textbf{All}} & \multicolumn{3}{c}{\textbf{EN}} \\
    \cmidrule(lr){2-4}\cmidrule(lr){5-7}
    \textbf{Model} & \textbf{False} & \textbf{True} & \textbf{Missing}
      & \textbf{False} & \textbf{True} & \textbf{Missing} \\
    \midrule
    Smart-Turn & \textbf{87.80} & \underline{98.81} & \textbf{95.35} & \textbf{88.82} & 99.00 & \textbf{95.35} \\
    FireRedChat & 75.24 & 9.29 & 51.19 & 72.23 & 14.34 & 51.19 \\
    Namo & 54.45 & 44.17 & 53.95 & 54.80 & 41.24 & 53.95 \\
    TEN & \underline{82.81} & 74.40 & \underline{79.08} & \underline{81.92} & 74.10 & \underline{79.08} \\
    TurnSense & 53.97 & 79.40 & 44.69 & 49.11 & 68.92 & 44.69 \\
    SoulX-Duplug & 38.76 & 94.52 & 56.32 & 35.60 & 96.02 & 56.32 \\
    X2-Turn & 41.53 & 98.57 & 67.08 & 40.54 & \textbf{99.60} & 67.08 \\
    \midrule
    Qwen3-0.6B & 75.12 & 0.00 & 48.87 & 72.41 & 0.00 & 48.87 \\
    XTurnix-PT & 30.47 & \textbf{98.93} & 70.41 & 29.92 & \underline{99.20} & 70.41 \\
    XTurnix-ZH-Base & 47.54 & 97.38 & 76.12 & 41.29 & 97.81 & 76.12 \\
    \bottomrule
  \end{tabular}
  \caption{Pooled (All) and English (EN) accuracy (\%) on Smart-Turn Bench
    grouped by \texttt{endfiller}, extending Table~\ref{tab:smartturn-endfiller}.
    False includes only explicitly false annotations; missing metadata is
    shown separately. Best results in each column
    are bold, and second-best results underlined.}
  \label{tab:smartturn-endfiller-all-en}
\end{table}

\begin{table}[htbp]
  \centering
  \small
  \setlength{\tabcolsep}{4pt}
  \begin{tabular}{@{}lllllllll@{}}
    \toprule
    & \multicolumn{4}{c}{\textbf{All}} & \multicolumn{4}{c}{\textbf{EN}} \\
    \cmidrule(lr){2-5}\cmidrule(lr){6-9}
    \textbf{Model} & \textbf{ROC-AUC} & \textbf{Start P} & \textbf{Start R} & \textbf{Keep R}
      & \textbf{ROC-AUC} & \textbf{Start P} & \textbf{Start R} & \textbf{Keep R} \\
    \midrule
    Qwen3-0.6B & 53.36 & 75.12 & 100.00 & 0.00 & 56.50 & 72.41 & 100.00 & 0.00 \\
    XTurnix-PT & 87.29 & 94.29 & 7.92 & 98.55 & 86.44 & 93.10 & 3.47 & 99.32 \\
    XTurnix-ZH-Base & 87.00 & 94.56 & 32.00 & 94.44 & 86.33 & 94.01 & 20.21 & 96.62 \\
    \bottomrule
  \end{tabular}
  \caption{Pooled (All) and English (EN) turn-completion metrics (\%)
    on Smart-Turn samples explicitly annotated as \texttt{endfiller=false},
    extending Table~\ref{tab:xturnix-start-conservatism}.}
  \label{tab:xturnix-start-conservatism-all-en}
\end{table}

\end{document}